\documentclass[11pt]{article}
\usepackage[preprint]{neurips_2026}
\usepackage[T1]{fontenc}
\usepackage[expansion=false]{microtype}
\usepackage{hyperref}
\usepackage{url}
\usepackage{booktabs}
\usepackage{amsfonts}
\usepackage{nicefrac}
\usepackage{xcolor}
\usepackage{graphicx}
\usepackage{tabularx}
\usepackage{colortbl}
\usepackage{makecell}
\usepackage{multirow}
\usepackage{float}
\usepackage{wrapfig}
\usepackage{enumitem}
\usepackage{pifont}
\usepackage{tikz}
\usepackage{algorithm}
\usepackage{algpseudocode}
\usetikzlibrary{shapes, arrows, positioning, fit, calc}
\usepackage{placeins}
\usepackage{needspace}
\usepackage{fvextra}

\usepackage[most]{tcolorbox}
\usepackage{xcolor}
\usepackage{graphicx}
\usepackage{fontawesome5}
\usepackage{enumitem}
\usepackage{caption}
\usepackage[normalem]{ulem} 
\newcommand{\best}[1]{\textbf{#1}}

\definecolor{lightblue}{RGB}{219,234,254}
\definecolor{lightgreen}{RGB}{220,252,231}
\definecolor{lightyellow}{RGB}{254,249,195}
\definecolor{lightred}{RGB}{254,226,226}
\definecolor{lightgray}{RGB}{243,244,246}
\definecolor{darkblue}{RGB}{30,64,175}
\definecolor{improve}{RGB}{22,163,74}
\definecolor{decline}{RGB}{220,38,38}
\definecolor{best}{RGB}{252,211,77}

\newcommand{\bench}{SVFSearch}
\newcommand{\cmark}{\ding{51}}   
\newcommand{\xmark}{\ding{55}}   

\title{\bench: A Multimodal Knowledge-Intensive Benchmark for Short-Video Frame Search in the Gaming Vertical Domain}

\author{%
  Lingtao Mao\thanks{Equal contribution.} \\
  Kuaishou Technology \\
  Beijing, China \\
  \texttt{mltzju@163.com} \\
  \And
  Huangyu Dai\footnotemark[1] \\
  Kuaishou Technology \\
  Beijing, China \\
  \texttt{11931034@zju.edu.cn} \\
  \And
  Xinyu Sun \footnotemark[1] \\
  Kuaishou Technology \\
  Beijing, China \\
  \texttt{sxy001122@gmail.com} \\
  \And
  Zihan Liang \\
  Kuaishou Technology \\
  Beijing, China \\
  \texttt{liangzih@seas.upenn.edu} \\
  \And
  Ben Chen\thanks{Corresponding author.} \\
  Kuaishou Technology \\
  Beijing, China \\
  \texttt{benchen4395@gmail.com} \\
  \AND
  Chenyi Lei \\
  Kuaishou Technology \\
  Beijing, China \\
  \texttt{leichy@mail.ustc.edu.cn} \\
  \And
  Wenwu Ou \\
  Kuaishou Technology \\
  Beijing, China \\
  \texttt{ouwenweu@gmail.com} \\
}

\date{}

\begin{document}

\maketitle

\begin{center}
\vspace{-0.8em}
\begin{tabular}{@{}l@{}}
\faGlobe~\textbf{Project Page:}~\href{https://svfsearch.github.io/SVFSearch-page/}{\texttt{https://svfsearch.github.io/SVFSearch-page/}} \\
\faGithub~\textbf{Code:}~\href{https://github.com/SVFSearch/SVFSearch-code}{\texttt{https://github.com/SVFSearch/SVFSearch-code}} \\
\faDatabase~\textbf{Dataset:}~\href{https://huggingface.co/datasets/svfsearch/SVFSearchData}{\texttt{https://huggingface.co/datasets/svfsearch/SVFSearchData}} \\
\end{tabular}
\vspace{0.3em}
\end{center}

\begin{abstract}
Multimodal large language models are increasingly used as agent backbones that understand multimodal inputs, plan retrieval actions, invoke external tools, and reason over retrieved information.
Yet existing benchmarks rarely evaluate this ability in short-video applications, where a paused frame is often visually ambiguous and answering requires vertical, long-tail, and fast-evolving domain knowledge.
We introduce SVFSearch, the first open benchmark for short-video frame search in the Chinese gaming domain. 
SVFSearch contains 5,000 four-choice test examples and 4,198 auxiliary training examples, each centered on a paused game scene from a real short-video clip.
To support fair and reproducible evaluation, SVFSearch provides a frozen offline retrieval environment with a game-domain text corpus, a topic-linked image gallery, and text, image, and multimodal retrieval interfaces, avoiding reliance on uncontrolled web search APIs.
We evaluate representative paradigms ranging from direct QA and RAG workflow to Plan-Act-Replan agents and learned search models.
Results reveal a large gap between model-only answering, practical agentic search, and oracle knowledge: the best open-source direct-QA model reaches 66.4\%, the best practical agent achieves 79.1\%, and oracle knowledge reaches 95.4\%.
Further analysis exposes bottlenecks in visual grounding, retrieval quality, evidence-grounded reasoning, and tool-use behavior, including over-search, answer-only shortcuts, and retrieval-induced misleading.
\end{abstract}

\section{Introduction}
\label{sec:intro}

Multimodal large language models (MLLMs) are shifting from passive multimodal predictors to active controllers in agentic systems, as seen in both proprietary systems such as GPT-5, Claude 4, and Gemini 3.1, and open-weight models such as Qwen3-VL and Qwen3.5~\citep{alayrac2022flamingo,li2023blip,liu2023visual,liu2024improved,wang2024qwen2, openai_gpt5_card,anthropic_claude4_card,google_gemini31_doc,bai2025qwen3}. 
They are increasingly used as agent backbones that analyze multimodal context, plan tool use, acquire external evidence, integrate returned information, and dynamically re-plan their reasoning process~\citep{yao2022react,shinn2023reflexion,langgraph}. 
Accordingly, recent benchmarks have begun to evaluate search-oriented multimodal behavior, including when to search, which tool to use, and how to process retrieved evidence to produce the final response~\citep{jiang2024mmsearch,tao2025mmsearch,li2024benchmarking,hu2024mrag}.

Despite this progress, existing benchmarks still miss a practical scenario that is increasingly common in short-video applications: \emph{search from a paused short-video frame}. When users pause on a salient frame, they often expect the system to explain the current visual context, retrieve background information, or answer follow-up questions. We refer to this setting as \emph{short-video frame search}. This scenario differs from standard image-only VQA because the visual query originates from a video, which may be accompanied by noisy video-side context, such as the video title, cover-frame OCR, and transcription. In this work, we focus the main evaluation on the paused frame and user question, then release the video-side metadata for future multi-source short-video research. Even under this controlled setting, answering the question often requires external domain knowledge that is vertical, long-tail, and fast-evolving. This makes short-video frame search distinct from encyclopedic visual QA~\citep{chen2023can,mensink2023encyclopedic} and existing multimodal-search benchmarks, which do not jointly study domain-specific visual evidence and text-based evidence retrieval.

In this work, we instantiate short-video frame search in the specialized Chinese gaming domain, where answering often requires recognizing game scenes, characters, equipment, maps, mechanics, version-specific content, and community knowledge. We introduce \bench, \textbf{the first open benchmark} for this setting. Each example is formulated as a four-choice QA task over a paused game frame and a corresponding user question, with the ground-truth answer and an answer rationale. \bench~contains 5,000 test items and 4,198 auxiliary train items. We additionally release video-side metadata, including the video title, cover-frame OCR, and ASR transcription, although these fields are not used in the main evaluation.

To support reproducible evaluation, we release an offline retrieval repository with a game-vertical text corpus and a topic-linked image gallery, avoiding dependence on online commercial search APIs. We evaluate representative paradigms on \bench, including direct QA, oracle knowledge QA, RAG workflow, LangGraph-based agents, and MMSearch-R1~\citep{langgraph,wu2025mmsearch}. 
Results reveal a clear search gap. On the same Qwen3.5-9B backbone, accuracy improves from 59.9\% with direct QA to 66.5\% with RAG workflow and further to 79.1\% with LangGraph-based agents, yielding a 19.2-point gain over the no-search setting. These findings highlight both the value of retrieval and the remaining challenges in evidence acquisition and efficient search.

Our contributions are summarized as follows:

\noindent(1) \textbf{We introduce \bench, the first agent-facing multimodal search benchmark for short-video frame search in the Chinese gaming vertical domain.} Built from real short-video clips, \bench~evaluates game-scene understanding and retrieval-augmented QA over paused frames and user questions, and releases video-side metadata for future multi-source evaluation.

\noindent(2) \textbf{We provide a reproducible offline retrieval environment.} \bench~ships a game-vertical text corpus, a topic-linked image gallery, and unified retrieval interfaces, enabling fair and reproducible evaluation without relying on uncontrolled online search APIs.

\noindent(3) \textbf{We benchmark representative multimodal-search paradigms and analyze their limitations.} We evaluate direct QA, oracle knowledge QA, RAG workflow, LangGraph-based agents, and MMSearch-R1, revealing a large performance gap and recurring tool-use failures such as over-search, answer-only shortcuts, and retrieval-induced misleading.

\vspace{-0.4em}
\section{\bench}
\vspace{-0.3em}
\label{sec:benchmark}

\begin{figure}[t]
  \centering
  \includegraphics[width=\textwidth]{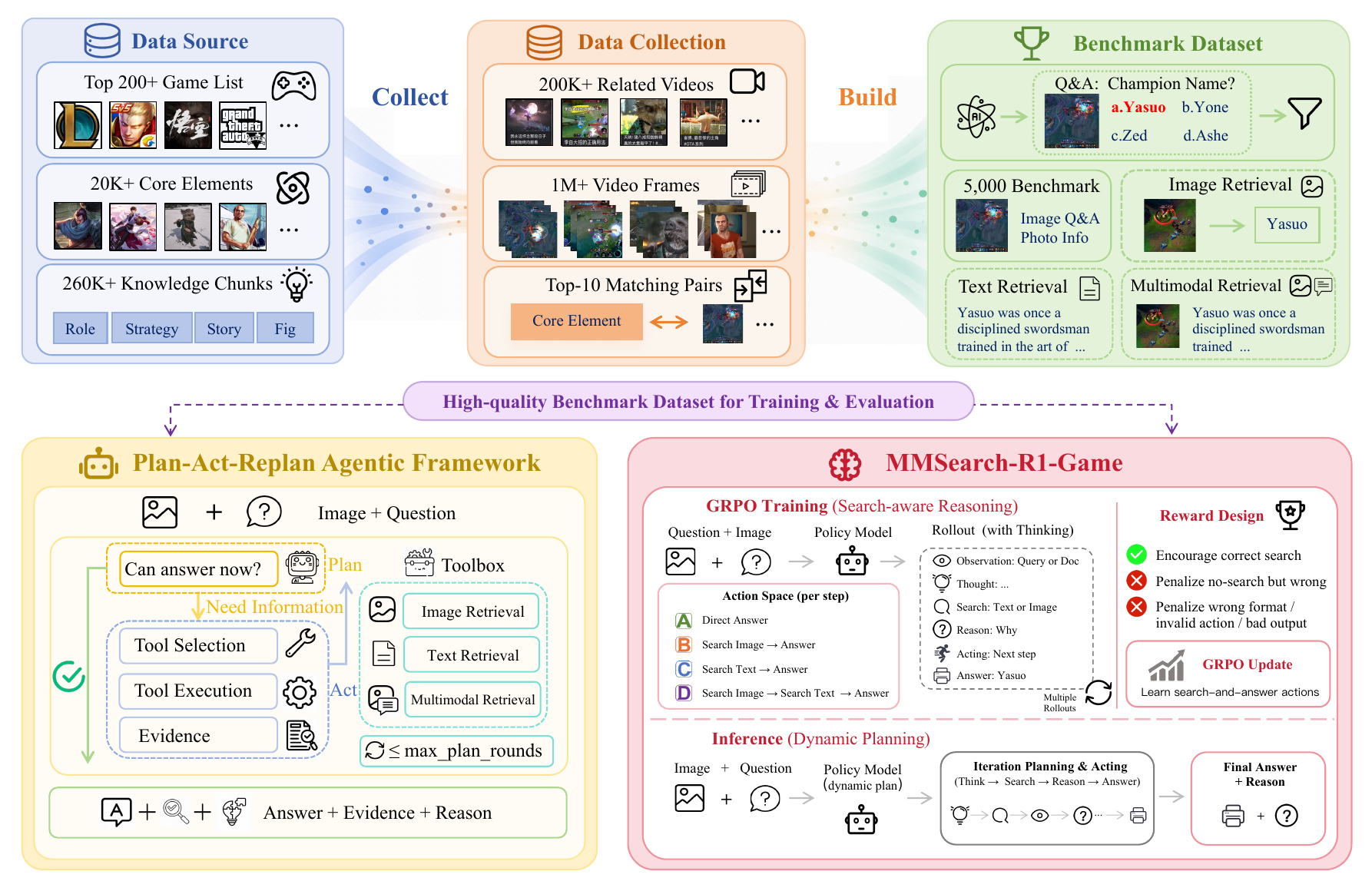}
    \caption{
      \textbf{Overview of \bench.}
      \emph{Top row:} benchmark construction from game-specific core elements,
      short-video frames, and web-sourced knowledge to QA splits and frozen
      retrieval resources.
      \emph{Bottom left:} a Plan-Act-Replan agent that dynamically decides whether more information is needed, selects retrieval tools, and integrates returned evidence before answering.
      \emph{Bottom right:} MMSearch-R1-Game, which learns search-and-answer actions through GRPO training in the same frozen retrieval environment.
    }
  \label{fig:overview}
\end{figure}

\subsection{Overview and Task Formulation}
\vspace{-0.3em}
\label{sec:benchmark_overview}

\bench~is a multimodal search benchmark with an agent-facing offline retrieval environment for game-scene understanding and retrieval-augmented question answering. Unlike standard visual QA benchmarks that mainly evaluate the direct answering ability of MLLMs, \bench~supports a broader range of systems, including direct-QA MLLMs, fixed RAG workflows, and autonomous multimodal agents that must decide whether to retrieve evidence and how to use it. Each example is centered on a paused game scene and formulated as a four-choice QA task that requires selecting the correct answer by grounding the visual context and using external evidence when needed.

\begin{figure}[t]
  \centering
  \includegraphics[width=0.78\linewidth]{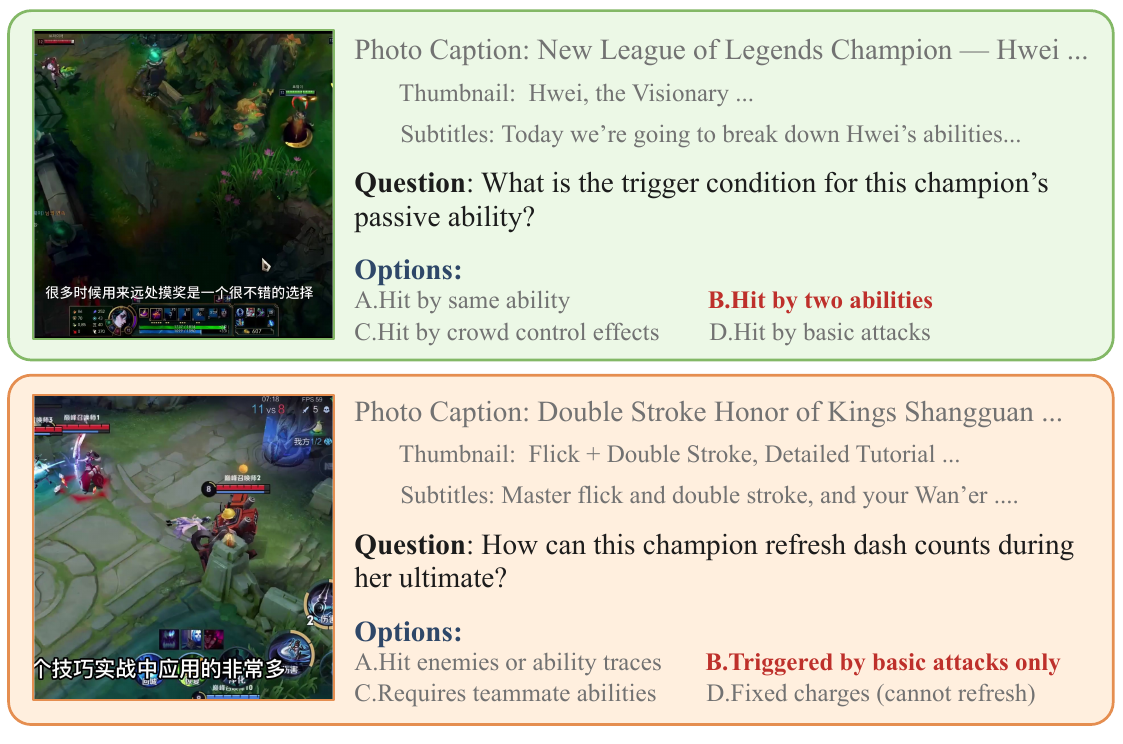}
  \caption{\textbf{Representative examples from \bench.}
  Examples show paused frames, video-side metadata, and multiple-choice QA instances.}
  \label{fig:case_examples}
  \vspace{-8pt}
\end{figure}

In \bench, each instance follows the format illustrated in Figure~\ref{fig:case_examples}: a paused-frame image, a question, four candidate answers, the ground-truth answer, and an answer rationale. We also release video-side metadata, including the video title, cover-frame OCR text, and ASR transcript, to support future studies of noisy video context and short-video information seeking. The benchmark contains a held-out test split and an auxiliary training split for training robust search-aware models. To support reproducible evaluation, \bench~further provides a standardized offline retrieval environment repository consisting of a game-vertical text corpus, a topic-linked image gallery, and frozen retrieval indices built over these resources, enabling controlled evaluation of visual grounding, tool selection, evidence retrieval, and final reasoning.

\subsection{Data Construction Pipeline}
\label{sec:construction}

As shown in Figure~\ref{fig:overview}, we construct \bench~around
\emph{game-specific core elements} rather than generic image captions or web QA
pairs. This design grounds each example in both concrete visual scenes and
game-domain knowledge. The pipeline has three stages: core-element and
knowledge construction, short-video-based visual grounding, and QA generation
with quality filtering.

\paragraph{Stage 1: Core Element and Knowledge Construction.}
We first collect 221 popular games covering diverse genres. Based on in-platform user queries, we mine game-specific \emph{core elements} for each game, including characters, equipment, maps, story events, skills, and gameplay mechanics. This process yields 22,800 core elements in total. For each core element, we retrieve multiple related knowledge sources through search engines. The retrieved content is then cleaned, summarized, and chunked with LLM assistance, producing standardized textual units for retrieval and QA construction. Finally, we obtain a text knowledge base with over 260K knowledge chunks grounded in the mined core elements.

\paragraph{Stage 2: Visual Grounding through Short-Video Retrieval.}
To connect textual game knowledge with concrete visual scenes, we use the game name and core element as the query to retrieve short-video content. In total, we collect more than 200K short videos related to the mined core elements, and extract over 1M candidate frames using \texttt{ffmpeg}. We then use an MLLM to verify whether each candidate frame visually matches the target core element, retaining up to 10 top-ranked frames for each core element. After filtering, we obtain 43,130 reliable core-element--image pairs with strong visual grounding. These pairs provide the visual basis for subsequent QA construction.

\paragraph{Stage 3: QA Generation and Quality Filtering.}
Given a visually grounded core element, its matched image, and the associated knowledge, an 8B-parameter model generates approximately 80K multiple-choice QA candidates that cover visual, knowledge-grounded, and reasoning-oriented questions. A 32B-parameter model then assesses each candidate's question quality, answer correctness, distractor plausibility, and difficulty. After quality assessment, difficulty annotation, and manual spot checks, we retain 9,198 high-quality QA instances, including 4,198 auxiliary training examples and 5,000 test examples. Each retained instance is linked back to its corresponding video-side metadata.

\subsection{Retrieval Resources and Frozen Indices}
\label{sec:retrieval_resources}

A central design goal of \bench~is to provide a reproducible offline retrieval environment rather than relying on online web search. We therefore release frozen retrieval resources for RAG workflows and multimodal agents. The image pool is built from the 43,130 visually grounded core-element--image pairs. We remove the 9,198 QA images to avoid exact-image retrieval leakage, resulting in 33,932 indexed game images, each associated with its game and core-element information. The text knowledge base contains 45,608 structured entries, segmented into 262,938 retrieval chunks. 
We build frozen retrieval resources over this corpus, including a dense text index using 512-dimensional Qwen3-Embedding-0.6B representations, a BM25 lexical retrieval tool, an image index using 256-dimensional features from a fine-tuned DINOv3-Base model, and a multimodal index using 512-dimensional Qwen3-VL-Embedding-2B representations for both image entries and text chunks.
Together, these resources enable controlled retrieval evaluation and help diagnose failures in visual grounding, entity matching, knowledge retrieval, evidence aggregation, and final reasoning.

\subsection{Benchmark Characteristics}
\label{sec:benchmark_characteristics}

\begin{table}[t]
\centering
\small
\caption{\textbf{Comparison with representative multimodal QA benchmarks.}
\cmark{} = present; \xmark{} = absent.
\textbf{T-KB}/\textbf{I-KB}: text/image retrieval resources.
\textbf{Agent}: supports tool-augmented retrieval evaluation.
\textbf{Dom.}: evaluation domain.
$^\dagger$ denotes an auxiliary training split for search-aware model development.}
\label{tab:comparison}
\setlength{\tabcolsep}{3.2pt}
\renewcommand{\arraystretch}{1.08}
\begin{tabular}{@{}lcccccc@{}}
\toprule
\textbf{Dataset} & \textbf{\#Test} & \textbf{\#Train} &
\textbf{T-KB} & \textbf{I-KB} & \textbf{Agent} &
\textbf{Dom.} \\
\midrule
VQAv2~\citep{goyal2017making}
& 214K & 443K & \xmark & \xmark & \xmark & General \\
OK-VQA~\citep{marino2019ok}
& 5.0K & 9.0K & \xmark & \xmark & \xmark & General \\
InfoSeek~\citep{chen2023can}
& 8.9K & 1.35M & Wiki & \xmark & \xmark & General \\
CharXiv~\citep{wang2024charxiv}
& 2.3K & -- & \xmark & \xmark & \xmark & Charts \\
GMAI-MMBench~\citep{chen2024gmai}
& 26K & -- & \xmark & \xmark & \xmark & Medical \\
Dyn-VQA~\citep{li2024benchmarking}
& 1.5K & -- & Web & Web & \cmark & General \\
MMSearch~\citep{jiang2024mmsearch}
& 300 & -- & Web & Web & \cmark & General \\
MMSearch-Plus~\citep{tao2025mmsearch}
& 311 & -- & Web & Web & \cmark & General \\
\midrule
\bench{} (ours)
& 5.0K & 4.2K$^\dagger$ & 45K & 34K & \cmark & Gaming \\
\bottomrule
\end{tabular}
\end{table}

\begin{figure}[t]
  \centering
  \includegraphics[width=\textwidth]{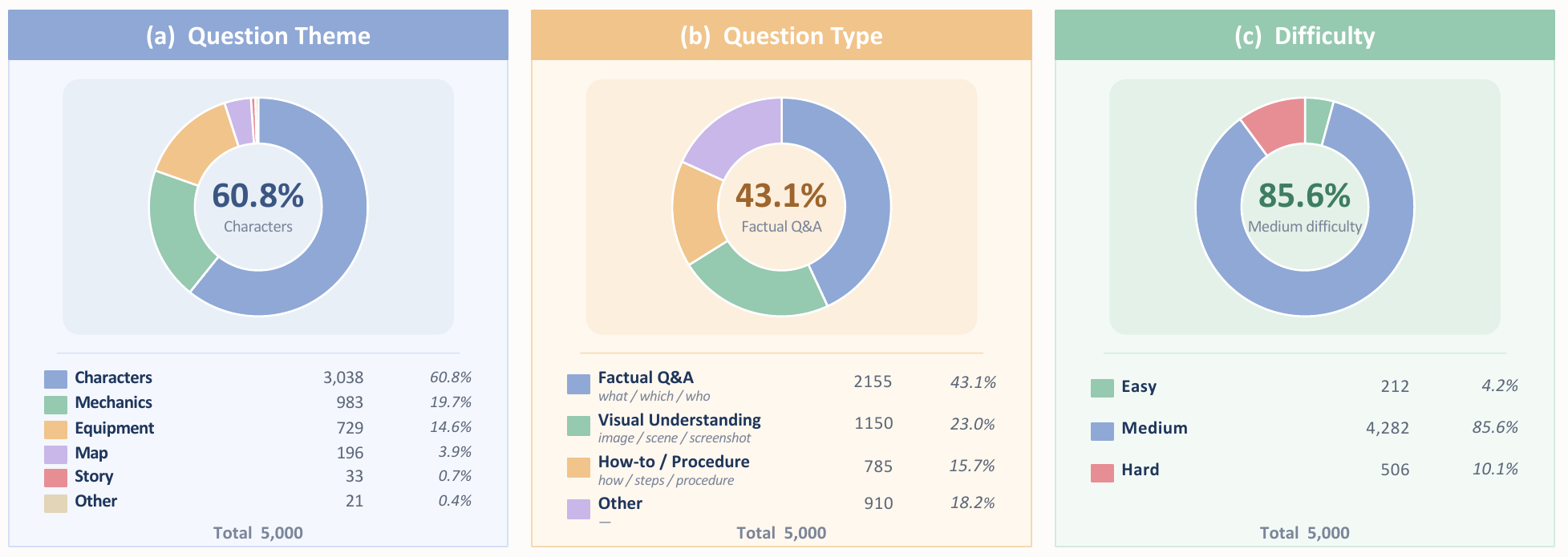}
    \caption{\textbf{Distribution analysis of \bench.}
    Test examples grouped by question theme, question type, and difficulty. The test split is dominated by character questions, factual Q\&A types, and medium-difficulty examples, while retaining long-tail themes and harder cases for stratified analysis.}
  \label{fig:distribution}
\end{figure}

As shown in Table~\ref{tab:comparison}, \bench~differs from prior multimodal QA benchmarks by coupling short-video frame questions with a frozen, domain-specific retrieval environment. The benchmark provides both text and image resources, enabling controlled evaluation of direct-QA, RAG workflows, and multimodal agents under the same offline setting. This design also allows agentic systems to decide whether additional evidence is needed and which retrieval interface to use, including text, image, and multimodal retrieval.

Figure~\ref{fig:distribution} shows that the 5,000-example test split reflects realistic game-scene search demand. Character and mechanics questions dominate, while equipment, map, story, and other long-tail themes are retained for stratified analysis. The benchmark further covers factual, visual-understanding, procedural, and reasoning-oriented questions across easy, medium, and hard difficulty levels. Each released instance has a unique image--question pair, while the same core element may appear in different visual contexts and support different questions.

\section{Evaluation Protocols and Search Baselines for \bench}
\label{sec:agentic_systems}

To evaluate representative paradigms on \bench, we build retrieval backends and search baselines using the frozen offline resources introduced in Section~\ref{sec:retrieval_resources}. We first present the core-element-aware image retrieval backend, and then instantiate the retrieval-dependent evaluation settings, including a fixed RAG workflow, a LangGraph-based Plan-Act-Replan agent, and MMSearch-R1 search models.

All tool-augmented settings operate on the same offline retrieval resources. The available tools include \texttt{img\_ann}, \texttt{text\_ann}, \texttt{bm25\_ann}, and \texttt{multimodal\_ann}. The outputs of \texttt{img\_ann} are adapted to different search settings. For MMSearch-R1 models, we keep \texttt{img\_ann} aligned with the original MMSearch-R1 image-retrieval interface, where an image search returns compact metadata rather than long knowledge passages. Specifically, each image entry in the ANN index is associated with metadata fields containing its game name and core element, so an \texttt{img\_ann} call returns concise element-level signals. For the LangGraph agent system, retrieved core elements from \texttt{img\_ann} are further used to invoke \texttt{kn\_lookup}, which returns detailed structured knowledge passages.

\subsection{Core-element-aware Image Retrieval Backend}
\label{sec:image_retrieval_training}

Image retrieval in \bench~is designed to identify the core element in a paused-frame image. We train a DINOv3-Base retrieval encoder with a cluster-aware contrastive strategy. For each core element, we first encode its associated images with DINOv3-Large and cluster the resulting features with K-means++, separating visually diverse appearances of the same element into multiple clusters.

During training, each epoch samples one positive pair per core element by randomly selecting one cluster and then sampling two images from that cluster. Images from other core elements in the mini-batch serve as negatives. We do not treat other clusters of the same element as negatives, since they may still represent the same core element under different appearances. At inference time, the fine-tuned DINOv3-Base performs nearest-neighbor retrieval over the frozen image index, and the retrieved images are mapped to core elements for downstream knowledge retrieval and answering.
  
\subsection{Fixed RAG Workflow and Plan-Act-Replan Agent}
\label{sec:langgraph_agent}

We include a fixed RAG workflow as the simplest tool-augmented baseline. It first calls \texttt{img\_ann} to retrieve visually similar images, aggregates the associated core elements by voting, and uses the selected element together with the original question to query \texttt{text\_ann}. The retrieved evidence is then provided to the MLLM along with the image, question, and options. Since the retrieval order is fixed, this workflow cannot adapt its search strategy based on intermediate evidence.

We further instantiate a Plan-Act-Replan (PAR) agent with LangGraph~\citep{langgraph}. At each round, the model observes the image, question, options, accumulated evidence, and previous tool outputs, then decides whether to answer or continue searching. When more evidence is needed, it selects one tool from \texttt{img\_ann}, \texttt{text\_ann}, \texttt{bm25\_ann}, and \texttt{multimodal\_ann}, and executes the corresponding tool call. The returned observation is appended to the evidence pool before the next planning round. We allow up to six rounds, with at most one tool call per round. The final answer is generated from the image, question, options, and accumulated evidence. Detailed algorithms for the fixed RAG workflow and PAR agent are provided in Appendix.

\subsection{MMSearch-R1 Reproduction and Game-domain Adaptation}
\label{sec:agentic_rl}

We evaluate \bench~with MMSearch-R1~\citep{wu2025mmsearch} (MS-R1) under three settings. First, we directly test the released MS-R1 model based on Qwen2.5-VL-7B. Second, we train \textbf{MS-R1-8B}, a Qwen3-VL-8B model using the MS-R1 codebase, our training data, and the original MS-R1 prompt and reward design. Third, we train \textbf{MS-R1-8B-Game}, which uses the same backbone and training framework but adapts the prompt and reward to the game-domain multiple-choice setting. Following the original MS-R1 protocol, all MS-R1-based models use only \texttt{text\_ann} and \texttt{img\_ann}.

Unlike PAR, MS-R1 internalizes tool use into generation: the model alternates between tool-call tokens and final-answer tokens, with retrieval observations appended to the context. For MS-R1-8B-Game, we adapt the reward to encourage explicit search behavior and reduce a multiple-choice shortcut where the model skips retrieval and guesses an option. Specifically, we penalize incorrect answers without search and add small rewards for valid image search and image--text search trajectories. Both MS-R1-8B variants are trained with GRPO following the MMSearch-R1 training framework. Detailed prompts are provided in Appendix.

\section{Experiments and Analysis}
\label{sec:experiments}

\begin{table*}[t]
\centering
\scriptsize
\setlength{\tabcolsep}{2.8pt}
\caption{\textbf{Comprehensive results on \bench.}
Accuracy and search rate (SR) are reported on the 5,000-example test split.
``---'' indicates settings without \bench{} retrieval tools.
$^\dagger$ denotes native thinking mode; Boldface indicates the best accuracy within each setting block.}
\label{tab:main_results}
\resizebox{\textwidth}{!}{%
\begin{tabular}{llccccccccccc}
\toprule
& & \multicolumn{1}{c}{\textbf{Overall}}
& \multicolumn{1}{c}{\textbf{Search}}
& \multicolumn{6}{c}{\textbf{Category}}
& \multicolumn{3}{c}{\textbf{Difficulty}} \\
\cmidrule(lr){3-3}\cmidrule(lr){4-4}\cmidrule(lr){5-10}\cmidrule(lr){11-13}
\textbf{Setting} & \textbf{Model}
& \textbf{Acc.} & \textbf{SR}
& \textbf{Char.} & \textbf{Equip.} & \textbf{Map} & \textbf{Story} & \textbf{Mech.} & \textbf{Other}
& \textbf{Easy} & \textbf{Med.} & \textbf{Hard} \\
\midrule

\multicolumn{13}{l}{\cellcolor{lightgray}\textit{Proprietary Models (Direct QA)}} \\
Direct & Claude-Opus-4.7 & 69.0 & --- & 70.0 & 69.0 & 66.8 & 72.7 & 66.2 & 66.7 & 67.9 & 68.6 & 72.9 \\
Direct & GPT-5.4 & 67.9 & --- & 69.8 & 67.8 & 59.7 & 69.7 & 63.7 & 66.7 & 65.1 & 67.7 & 70.4 \\
Direct & \best{Gemini-3.1-Pro} & \best{77.5} & --- & 79.1 & 79.6 & 69.9 & 87.9 & 72.3 & 71.4 & 82.5 & 78.3 & 68.2 \\
\midrule

\multicolumn{13}{l}{\cellcolor{lightgray}\textit{Open-source Direct QA}} \\
Direct & Qwen2.5-VL-7B & 49.8 & --- & 50.8 & 47.9 & 47.4 & 51.5 & 48.1 & 57.1 & 57.1 & 49.4 & 50.2 \\
Direct & Qwen2.5-VL-7B-CoT & 47.4 & --- & 47.8 & 47.6 & 46.4 & 60.6 & 45.8 & 42.9 & 64.6 & 47.5 & 39.3 \\
Direct & Qwen3-VL-8B & 54.8 & --- & 56.5 & 52.4 & 55.1 & 66.7 & 50.9 & 47.6 & 58.0 & 53.8 & 61.9 \\
Direct & Qwen3-VL-8B-Thinking  & 53.7 & --- & 54.1 & 54.0 & 52.0 & 60.6 & 52.0 & 61.9 & 74.1 & 52.8 & 52.8 \\
Direct & Qwen3-VL-32B & 57.1 & --- & 58.7 & 54.3 & 53.6 & 63.6 & 54.4 & 71.4 & 71.2 & 56.2 & 58.9 \\
Direct & Qwen3-VL-32B-Thinking  & 62.2 & --- & 63.4 & 60.9 & 58.2 & 69.7 & 59.8 & 76.2 & 75.5 & 61.9 & 59.9 \\
Direct & Qwen3.5-9B & 59.9 & --- & 61.1 & 57.1 & 57.1 & 60.6 & 58.6 & 66.7 & 59.9 & 59.2 & 65.4 \\
Direct & Qwen3.5-9B$^{\dagger}$ & 58.3 & --- & 59.3 & 55.7 & 61.2 & 66.7 & 56.6 & 57.1 & 63.7 & 57.6 & 62.3 \\
Direct & \best{Qwen3.5-27B} & \best{66.4} & --- & 68.0 & 63.1 & 63.8 & 75.8 & 63.9 & 76.2 & 73.6 & 65.5 & 71.3 \\
Direct & Qwen3.5-27B$^{\dagger}$ & 64.8 & --- & 65.4 & 62.5 & 61.2 & 72.7 & 64.8 & 66.7 & 69.8 & 64.8 & 62.5 \\
\midrule

\multicolumn{13}{l}{\cellcolor{lightgray}\textit{Open-source RAG Workflow}} \\
Workflow & Qwen2.5-VL-7B & 57.3 & 100.0 & 57.7 & 58.6 & 58.2 & 69.7 & 54.3 & 52.4 & 59.4 & 57.9 & 51.0 \\
Workflow & Qwen3-VL-8B & 63.5 & 100.0 & 64.1 & 66.5 & 63.3 & 60.6 & 59.6 & 61.9 & 66.5 & 64.3 & 55.3 \\
Workflow & Qwen3-VL-32B & 62.2 & 100.0 & 61.6 & 67.2 & 62.8 & 69.7 & 60.3 & 61.9 & 78.3 & 62.7 & 51.4 \\
Workflow & Qwen3.5-9B & 66.5 & 100.0 & 67.5 & 67.2 & 64.3 & 63.6 & 63.3 & 61.9 & 70.8 & 66.5 & 64.4 \\
Workflow & \best{Qwen3.5-27B} & \best{69.4} & 100.0 & 70.8 & 69.4 & 65.8 & 78.8 & 65.3 & 66.7 & 70.8 & 69.1 & 71.0 \\
\midrule

\multicolumn{13}{l}{\cellcolor{lightgray}\textit{Open-source Plan-Act-Replan Agent}} \\
PAR & Qwen2.5-VL-7B & 59.3 & 85.6 & 59.3 & 61.7 & 63.3 & 54.5 & 56.9 & 52.4 & 55.2 & 59.8 & 56.5 \\
PAR & Qwen3-VL-8B & 63.7 & 99.7 & 62.8 & 66.1 & 70.9 & 63.6 & 63.2 & 66.7 & 66.5 & 64.2 & 58.3 \\
PAR & Qwen3-VL-32B & 71.6 & 98.4 & 71.1 & 75.9 & 74.5 & 69.7 & 69.4 & 76.2 & 79.7 & 72.4 & 61.5 \\
PAR & \best{Qwen3.5-9B} & \best{79.1} & 100.0 & 79.3 & 82.7 & 79.1 & 87.9 & 76.0 & 71.4 & 75.0 & 79.7 & 75.9 \\
PAR & Qwen3.5-27B & 78.6 & 96.8 & 78.3 & 83.1 & 81.1 & 81.8 & 75.5 & 81.0 & 79.7 & 79.0 & 74.5 \\
\midrule

\multicolumn{13}{l}{\cellcolor{lightgray}\textit{Open-source MMSearch-R1 Search Models}} \\
MS-R1 & Qwen2.5-VL-7B & 49.4 & 72.8 & 49.1 & 53.8 & 54.1 & 60.6 & 46.2 & 33.3 & 49.1 & 35.5 & 29.8 \\
MS-R1 & Qwen3-VL-8B & 63.2 & 0.02 & 64.5 & 60.8 & 63.3 & 63.6 & 61.1 & 52.4 & 70.3 & 62.6 & 65.6 \\
MS-R1-Game & \best{Qwen3-VL-8B} & \best{64.5} & 68.2 & 65.2 & 66.8 & 65.3 & 63.6 & 60.0 & 76.2 & 73.6 & 64.3 & 61.9 \\
\midrule

\multicolumn{13}{l}{\cellcolor{lightgray}\textit{Open-source Oracle Knowledge (Upper Bound)}} \\
Oracle & Qwen3-VL-8B & 86.5 & --- & 88.4 & 87.5 & 87.8 & 78.8 & 80.3 & 76.2 & 73.1 & 87.5 & 84.0 \\
Oracle & Qwen3-VL-32B & 90.8 & --- & 92.0 & 94.1 & 88.3 & 93.9 & 85.4 & 81.0 & 87.7 & 91.4 & 87.0 \\
Oracle & Qwen3.5-9B & 90.3 & --- & 90.7 & 91.9 & 86.2 & 97.0 & 88.4 & 85.7 & 91.5 & 90.7 & 85.8 \\
Oracle & \best{Qwen3.5-27B} & \best{95.4} & --- & 96.0 & 96.7 & 95.9 & 87.9 & 93.0 & 90.5 & 94.3 & 95.7 & 93.5 \\
\bottomrule
\end{tabular}}
\vspace{-1em}
\end{table*}

\subsection{Experimental Setup}
\label{sec:experimental_setup}

We evaluate all methods on the 5,000-example test split of \bench, using accuracy as the main metric. Unless otherwise specified, each method receives only the paused-frame image, question, and answer options. The released video title, cover-frame OCR text, and ASR transcript are excluded from the main experiments.

We compare five evaluation settings. Direct QA evaluates MLLMs without retrieval tools. RAG Workflow follows the fixed image-to-text retrieval workflow described in Section~\ref{sec:langgraph_agent}. PAR evaluates the LangGraph-based Plan-Act-Replan agent from the same section. MS-R1 evaluates MMSearch-R1-style models that generate tool calls as part of the decoding process. Oracle Knowledge provides the ground-truth knowledge associated with each question and serves as an upper bound on evidence availability. All retrieval-augmented settings use the same frozen retrieval resources to ensure fair comparison. RAG Workflow and PAR use top-5 retrieval, while MS-R1 uses top-3 image retrieval and top-5 text retrieval. We also report search rate, denoted as SR, which measures the fraction of examples where a method invokes at least one retrieval tool. Direct QA and Oracle Knowledge do not invoke \bench~retrieval tools, so their SR is marked as ``---''. Qwen2.5-VL-7B-CoT uses prompt-induced step-by-step reasoning, while Thinking models use their native thinking mode. Further implementation details are provided in Appendix.

\subsection{Main Results}
\label{sec:main_results}

Table~\ref{tab:main_results} reports results on the 5,000-example test split. Full experimental results and additional analyses are provided in Appendix. Overall, \bench~is challenging when methods rely only on direct MLLM answering. Among proprietary models, the strongest Direct QA result is achieved by Gemini-3.1-Pro at 77.5\%, while the best open-source Direct QA setting reaches 66.4\% with Qwen3.5-27B. In contrast, Oracle Knowledge substantially improves performance, reaching 95.4\% with Qwen3.5-27B and 90.8\% with Qwen3-VL-32B. This gap shows that many questions are answerable when the relevant game-specific evidence is available, but remain difficult without retrieved evidence.

\begin{figure}[t]
  \centering
  \includegraphics[width=\textwidth]{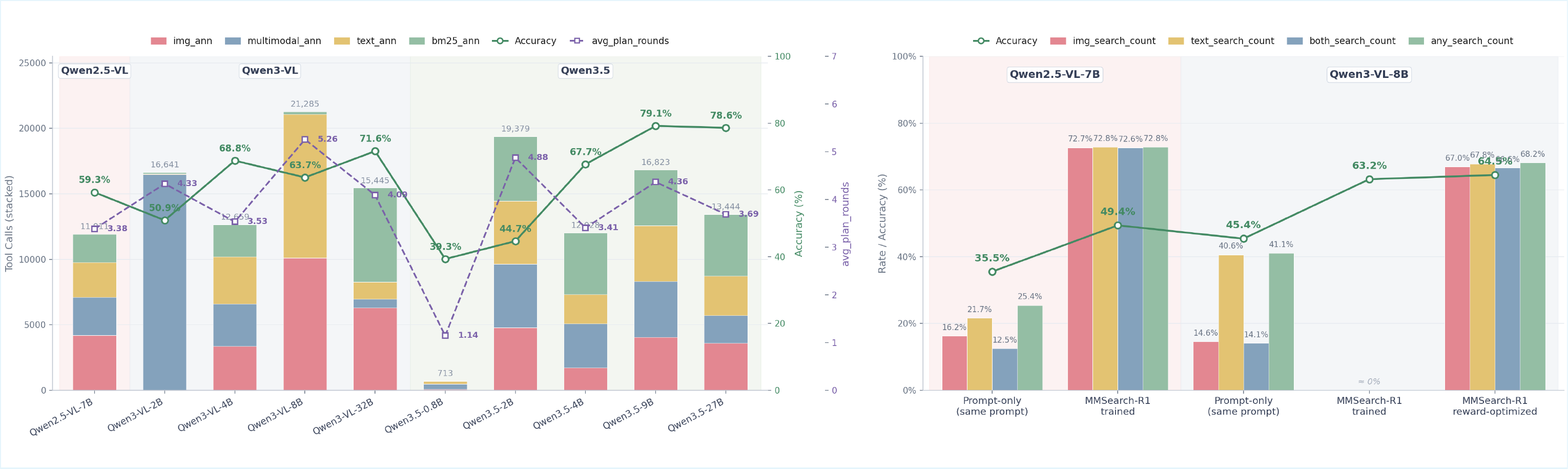}
    \caption{\textbf{Tool-use diagnostics.}
    Left: PAR tool calls, accuracy, and average planning rounds across backbones.
    Right: item-level search rates and accuracy of MS-R1-style models.
    Search-rate bars on the right are not mutually exclusive.}
  \label{fig:tool_diagnostics}
  \vspace{-1.5em}
\end{figure}
  
Retrieval improves open-source methods across most settings. With Qwen3-VL-8B as the backbone, accuracy increases from 54.8\% under Direct QA to 63.5\% with RAG Workflow, 63.7\% with PAR, and 64.5\% with MS-R1-Game. For stronger backbones, PAR gives the best practical results: Qwen3.5-9B with PAR reaches 79.1\%, surpassing all Direct QA baselines, including Gemini-3.1-Pro. The high SR values under PAR, ranging from 85.6\% to 100.0\%, show that these agents frequently rely on retrieval, and the corresponding gains indicate that external evidence is often beneficial. Model scaling and version upgrades are most visible on easy examples: under Direct QA, Qwen3-VL improves from 58.0\% at 8B to 71.2\% at 32B, and Qwen3.5-27B reaches 73.6\%. Hard examples still remain far below Oracle Knowledge, suggesting persistent bottlenecks in visual grounding, retrieval quality, evidence selection, and evidence-grounded reasoning.

The MS-R1 results highlight both the promise and the difficulty of learned tool use. MS-R1-Game improves Qwen3-VL-8B to 64.5\% with an SR of 68.2\%, showing that task-adapted training can recover meaningful retrieval behavior while improving answer accuracy. By contrast, the original MS-R1 training improves Qwen3-VL-8B to 63.2\% but yields an SR of only 0.02\%, indicating that its gain mainly comes from improved reasoning behavior rather than actual retrieval use. This suggests a limitation of outcome-only RL in multiple-choice settings: a model can improve answer accuracy while learning to bypass search, because guessing can still receive a non-trivial reward.  Effective MS-R1 training on \bench~therefore requires task-specific reward design that encourages retrieval when evidence is needed while preserving answer correctness.

\subsection{Tool-use Diagnostics}
\label{sec:tool_use_diagnostics}

\paragraph{PAR tool-use behavior.}
The left panel of Figure~\ref{fig:tool_diagnostics} shows that PAR behavior varies substantially across model scales. Qwen3.5-0.8B makes only 713 tool calls, with an average of 1.14 planning rounds, and achieves 39.3\%, suggesting that very small models struggle to sustain multi-step search behavior. Qwen3.5-2B shows the opposite pattern: it makes 19,379 tool calls over 4.88 rounds but reaches only 44.7\%, indicating excessive search without effective use of evidence. In contrast, Qwen3.5-27B makes fewer calls, 13,444 with 3.69 rounds, yet reaches 78.6\%. These results show that PAR performance is not determined by the number of tool calls alone, but also depends on evidence selection and integration.

Tool preference also varies across model families. Qwen3-VL models show uneven tool-use patterns across scales, with different sizes favoring different retrieval channels, while Qwen3.5 models above 4B use multiple channels more consistently. The best PAR setting, Qwen3.5-9B, reaches 79.1\% with relatively balanced use of image, text, BM25, and multimodal retrieval. This suggests that effective PAR benefits from coordinating complementary retrieval channels rather than simply increasing the number of calls.

\paragraph{MS-R1 tool-use behavior.}
The right panel of Figure~\ref{fig:tool_diagnostics} compares prompt-only and RL-trained MS-R1 variants. Prompt-only tool use is weak: under the same search prompt, Qwen2.5-VL-7B achieves only 35.5\% accuracy with a 25.4\% any-search rate, and Qwen3-VL-8B achieves 45.4\% accuracy with a 41.1\% any-search rate. Both are below their Direct QA counterparts in Table~\ref{tab:main_results}, showing that long tool-use prompts alone can introduce instability and hurt performance.

\begin{figure*}[!htbp]
  \centering
  \includegraphics[width=\textwidth]{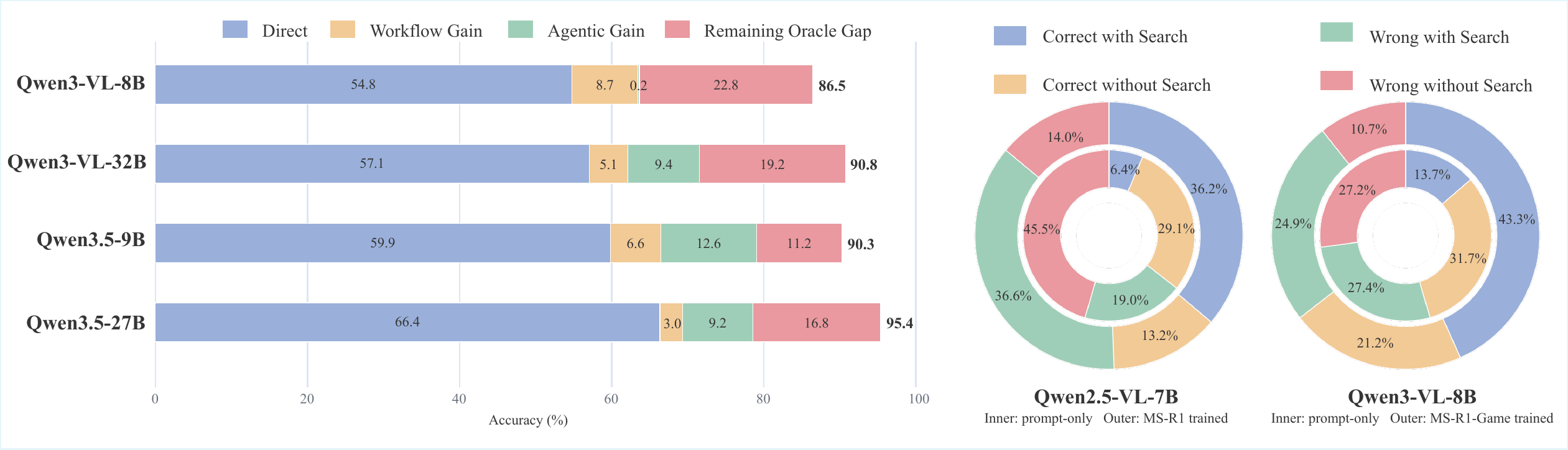}
    \caption{\textbf{Retrieval gains and search behavior.}
    Left: accuracy decomposition from Direct QA to RAG Workflow, PAR, and Oracle Knowledge.
    Right: correctness and search-usage breakdown for prompt-only and trained MS-R1-style models.}
  \label{fig:gain_and_behavior}
  \vspace{-1.0em}
\end{figure*}

RL training changes tool-use behavior, but the effect depends strongly on the task and reward design. The released Qwen2.5-VL-7B MMSearch-R1 model searches on 72.8\% of examples, yet remains slightly below its Direct QA baseline, showing that a trained search policy does not transfer automatically to the game-domain QA setting. For Qwen3-VL-8B, the original MS-R1 training improves accuracy to 63.2\% but almost eliminates tool use, indicating that the model mainly learns stronger reasoning behavior rather than reliable search. Our reward-adapted MS-R1-Game restores meaningful retrieval, reaching 64.5\% accuracy with a 68.2\% any-search rate and frequent use of both image and text search. These results support the value of agentic-search RL, while showing that task-specific data and reward design are needed to avoid answer-only shortcuts in multiple-choice settings.

\subsection{Retrieval Gains and Search Behavior}
\label{sec:gain_and_behavior}

Figure~\ref{fig:gain_and_behavior} summarizes how retrieval changes model performance and search behavior. The left panel decomposes the accuracy from Direct QA to Oracle Knowledge. Across Qwen3-VL and Qwen3.5 backbones, RAG Workflow consistently improves over Direct QA, and PAR can add further gains by allowing adaptive tool use beyond a fixed retrieval chain. For example, PAR adds 9.4 points over RAG Workflow on Qwen3-VL-32B and 12.6 points on Qwen3.5-9B. However, the remaining gap to Oracle Knowledge is still large, including 19.2 points for Qwen3-VL-32B and 16.8 points for Qwen3.5-27B, indicating persistent bottlenecks in visual grounding, query formulation, retrieval quality, and evidence-grounded reasoning.

The right panel compares prompt-only and trained MS-R1 models by correctness and search usage. After training, correct-with-search examples increase from 6.4\% to 36.2\% for Qwen2.5-VL-7B and from 13.7\% to 43.3\% for Qwen3-VL-8B, while wrong-without-search examples decrease from 45.5\% to 14.0\% and from 27.2\% to 10.7\%, respectively. This shows that training makes search more useful in evidence-demanding cases. At the same time, correct-without-search examples decrease and wrong-with-search examples remain non-negligible, suggesting that learned search policies can still over-search or retrieve unhelpful evidence. These results support learned tool use on \bench, but also show the need for reward and data design that balances search necessity, search quality, and answer correctness.

\section{Limitations}
\label{sec:limitations}

\bench~focuses on short-video frame search in the Chinese gaming domain, which provides a controlled and realistic setting for evaluating domain-specific multimodal search. This focus also means that the results should be interpreted within this vertical domain rather than as a universal estimate of all short-video applications. In addition, our main evaluation uses four-choice QA and a frozen offline retrieval environment to ensure stable and reproducible comparisons. Although we release video-side metadata such as titles, cover-frame OCR text, and ASR transcripts, their full use is left to future multi-source evaluation. Future work can extend the benchmark to open-ended answering, additional vertical domains, and evolving retrieval corpora.

\section{Conclusion}
\label{sec:conclusion}

We introduced \bench, a benchmark for short-video frame search in the Chinese gaming domain, together with a frozen offline retrieval environment spanning text, image, and multimodal evidence. The benchmark supports controlled evaluation of direct-QA MLLMs, fixed RAG workflows, planner-based agents, and MS-R1 learned search models under the same retrieval resources. Experiments show that game-specific evidence is critical for many examples, while practical agents still face substantial gaps in visual grounding, tool-use control, retrieval quality, and evidence integration. We hope \bench~can serve as a practical testbed for studying retrieval-augmented and agentic multimodal search over paused frames from short videos.

\bibliographystyle{plainnat}
\bibliography{neurips_reference}







\end{document}